# Jira: a Kurdish Speech Recognition System Designing and Building Speech Corpus and Pronunciation Lexicon

Hadi Veisi*
University of Tehran,
Faculty of New Sciences and Technologies
h.veisi@ut.ac.ir

Hawre Hosseini
Ryerson University,
Electrical and Computer Engineering
hawre.hosseini@ryerson.ca

Mohammad MohammadAmini
Avignon University,
Laboratoire Informatique d'Avignon (LIA)
mohammad.mohammadamini@univ-avignon.fr

Wirya Fathy
University of Tehran,
Faculty of New Sciences and Technologies
w.fathy@ut.ac.ir

Aso Mahmudi
University of Tehran,
Faculty of New Sciences and Technologies
aso.mahmudi@ut.ac.ir

**Abstract:** In this paper, we introduce the first large vocabulary speech recognition system (LVSR) for the Central Kurdish language, named Jira. The Kurdish language is an Indo-European language spoken by more than 30 million people in several countries, but due to the lack of speech and text resources, there is no speech recognition system for this language. To fill this gap, we introduce the first speech corpus and pronunciation lexicon for the Kurdish language. Regarding speech corpus, we designed a sentence collection in which the ratio of di-phones in the collection resembles the real data of the Central Kurdish language. The designed sentences are uttered by 576 speakers in a controlled environment with noise-free microphones (called AsoSoft Speech-Office) and in Telegram social network environment using mobile phones (denoted as AsoSoft Speech-Crowdsourcing), resulted in 43.68 hours of speech. Besides, a test set including 11 different document topics is designed and recorded in two corresponding speech conditions (i.e., Office and Crowdsourcing). Furthermore, a 60K pronunciation lexicon is prepared in this research in which we faced several challenges and proposed solutions for them. The Kurdish language has several dialects and sub-dialects that results in many lexical variations. Our methods for script standardization of lexical variations and automatic pronunciation of the lexicon tokens are presented in detail. To set-up the recognition engine, we used the Kaldi toolkit. A statistical tri-gram language model that is extracted from the AsoSoft text corpus is used in the system. Several standard recipes including HMM-based models (i.e., mono, tri1, tr2, tri2, tri3), SGMM, and DNN methods are used to generate the acoustic model. These methods are trained with AsoSoft Speech-Office and AsoSoft Speech-Crowdsourcing and a combination of them. The best performance achieved by the SGMM acoustic model which results in 13.9% of the average word error rate (on different document topics) and 4.9% for the general topic.

## 1 Introduction

Nowadays, speech technologies and mainly automatic speech recognition (ASR) are available in many real applications such as voice dictation systems, smart assistants, telephone information systems, command and control, games and entertainments, and speech translators. There is about seven decades' history between the early endeavors for developing a small vocabulary speaker-dependent ASR system in the 1950s from Bell labs (Furui 2005) until now when very large vocabulary speaker-independent systems are released in real applications by large companies such as Google, Microsoft, Apple, and Amazon. Methods used in ASR systems have also progressed along with this history from dynamic time warping (DTW) algorithm (Sakoe and Chiba 1978; Vintsyuk 1968) to hidden Markov model (HMM) (Baum 1972; Jelinek 1976; Levinson et al. 1983) and then deep feedforward and recurrent neural networks (Hinton et al. 2012; Yu and Deng 2016). In parallel, the rapid growth of available data for languages and also improvement of the hardware requirements for processing the large-scale data have provided the necessary conditions for developing high-quality ASR systems. As a result, today speech recognition systems are available for most languages and easily accessible on smartphones. Despite the remarkable progress in speech recognition development, it is not surprising that there are languages with many speakers such as Kurdish, for which such systems have not been developed.

Kurdish as an Indo-Iranian language, is spoken by more than 30 million people in western Asia mainly in Turkey, Iraq, Iran, Syria, Armenia, and Azerbaijan. The Kurdish language has various dialects categorized in three main branches including Central Kurdish (also called Sorani), Northern Kurdish (also called Kurmanji), and Southern Kurdish. Although this language is spoken by a large population, it suffers from the unavailability of sufficient language resources for its computational processing purposes. The lack of language resources such as speech and text corpora seems to be the main reason for the lack of development of speech recognition and other natural language processing (NLP) applications in Kurdish. To develop a practical ASR system, three main linguistic resources are required: a speech corpus for acoustic modeling, a text corpus for language modeling and a pronunciation lexicon. Although today there are integrated approaches such as end-to-end speech recognition (Amodei et al. 2016; Graves and Jaitly 2014; He et al. 2019) that mainly utilize a large speech corpus to learn all information about the language, the most practical speech recognition systems still are leveraging separate components and separate data sources for the learning (Huang et al. 2001; Yu and Deng 2016).

To the best of our knowledge, there was not any Kurdish speech corpus for speech recognition purposes before our endeavor. There are limited attempts toward developing a speech-related technology for Kurdish, among them developing text-to-speech (TTS) systems (Barkhoda et al. 2009), dialect recognition (Al-Talabani et al. 2017), speaker recognition (Abdul 2019) and digit recognition (Abdulrahman et al. 2019). The speech datasets developed during those researches are very limited and of course, are not applicable to real-time speech recognition applications.

However, there are numbers of research for developing computational resources of the Kurdish language. In (Haig 2002), a small corpus called *corpus of contemporary Kurdish newspaper texts (CCKNT)* is collected which includes 214K words of the Northern Kurdish dialect. In (Sheykh Esmaili and Salavati 2013), *Pewan* text corpus is introduced for Central Kurdish and Northern Kurdish which was collected from two online news agencies. The size of the Pewan corpus for the Central Kurdish dialect is about 18M tokens and for Northern Kurdish dialect is about 4M. This corpus is used as a test set for information retrieval applications (Sheykh Esmaili et al. 2013). The largest text corpus for Central Kurdish was introduced by AsoSoft team (Veisi et al. 2019) called *AsoSoft Text Corpus*. This corpus includes about 188M tokens and has been collected from different sources mainly from Kurdish websites, published books, magazines, and newspapers. Among the aforementioned text corpora, AsoSoft Text Corpus is the only big enough resource that can be used in developing a speech recognition system.

There are also limited works for the development of Kurdish computational lexica. Two similar works have been done by Walther (Walther et al. 2010; Walther and Sagot 2010) to build the Kurdish lexicon automatically. In (Walther and Sagot 2010), a three-step semi-supervised method is proposed for developing a morphological lexicon for less-resourced languages and is applied for Central Kurdish to extract a lexicon from a small corpus of size 590K tokens. In (Walther et al. 2010), similar research has been done for Northern Kurdish. In another research in (Hosseini et al. 2015), authors have proposed a method to build a generative lexicon of Central Kurdish, including 35K tokens. Although there are several dictionaries, they cannot be used alone for extracting a lexicon of speech recognition since they do not include most forms of the words and do not contain all of the most frequent today's words.

In this paper, we introduce the first ASR system for Central Kurdish. To do this, we review our works on designing and collecting a speech corpus for this language. The proposed speech corpus is also the first large-scale speech corpus (more than 43 hours) that can be utilized in the development of ASR systems as well as other speech processing tasks such as speaker recognition. Furthermore, we have developed a pronunciation lexicon (of size 60K) to be used in the ASR system and other related NLP tasks such as TTS and data-driven grapheme-to-phoneme (G2P). The implementation of an ASR engine for Central Kurdish and the evaluation of this system are also presented. For the sake of readability, in the continuant, we use Kurdish instead of Central Kurdish (Sorani) where there is no need to make a distinction between Kurdish dialects.

In this paper, we first overview the Kurdish language focusing on the phonology of the Central Kurdish dialect in Section 2. In Section 3, our experiments in designing and collecting the AsoSoft Speech Corpus are described. The development of the first Kurdish pronunciation lexicon for speech recognition is presented in Section 4. In Section 5, we introduce Jira, the first Kurdish speech recognition system. In this section, we provide the system architecture, acoustic model (AM) and language model (LM) details and results of the evaluations. Finally, the conclusion and future works are given in Section 6.

## 2 Kurdish Language

In this section, we elaborate on different dialects of the Kurdish language. Moreover, we give a more detailed explanation of Central Kurdish writing system and some important pronunciation points. The aforementioned explanations are important since they have determined some of the automatizations for building lexicon in this work. The Kurdish language is spoken by various dialects that are categorized into three groups known as:

1. Central Kurdish (Sorani) is spoken by the majority of Kurds residing in Iraq and Iran. The written standard form of Central Kurdish is called Sorani and was developed in the 1920s (Nebez 1993) and was later adopted as the standard orthography of Kurdish as an official language in Iraq.
2. Northern Kurdish (Kurmanji) is spoken in the Northern areas of Kurdistan including Turkey, Northern Iraq, Northwestern Iran, and Northern Syria.
3. Southern Kurdish is the language of Kurds in Kermanshah and Ilam provinces of Iran and Southern part of Iraqi Kurdistan (e.g., Khanaqin district).

Other dialects spoken by smaller populations are Zazaki and Gorani (also known as Hawrami).

The two major dialects of Kurdish are Northern Kurdish (Kurmanji) and Central Kurdish (Sorani). Northern Kurdish is written in Latin (Roman) script while the Central Kurdish dialect is mostly written in a customized version of the Arabic script. Although Northern Kurdish has a greater number of speakers than the Central Kurdish, the latter has more written resources.

The Arabic-based writing system of Central Kurdish was first established in the 1920's (Nebez 1993) and has undergone drastic changes ever since. Table 1 shows the phonological features of the consonants and vowels of the language, compared to 34 letters of its alphabet with an example for each one. This writing system is almost phonemic, i.e., each letter corresponds to one phoneme, with a few exceptions. As it is given in rows 28 and 30, the letter 'ی' is both pronounced as /j/ (palatal approximant) and as /i/ (a vowel). Also, as shown in rows 29 and 34, the letter ‘و’ is pronounced both as /w/ (bilabial approximant) and as /ʊ/ (a vowel). This letter is also written in a repeated form as ‘وو’ and is pronounced as /u/ (a long vowel). Although the digraph ‘وو’ represent one phoneme, it is not considered as an independent character in the keyboard layouts released for Kurdish by the department of information technology of Kurdistan Regional Government (2014). Besides, there is a short vowel /i/ in Kurdish (row 37 in Table 1) which is not written in the Arabic script (e.g., in the word “دڵ” [heart], /i/ is a phoneme between “د” and “ڵ”) but is written in the Latin script of Kurdish.

Table 1: Central Kurdish phonemes and letters

| **No** | **Feature** | | | **IPA** | **Phoneme** | **Letter (isolated form)** | **Example** |
|---|---|---|---|---|---|---|---|
| **1** | Consonants | Voiced | Stop | b | b | ب | باوان |
| **2** | | | | d | d | د | کورد |
| **3** | | | | d͡ʒ | je | ج | گۆج |
| **4** | | | | g | g | گ | درەنگ |
| **5** | | | Fricative | v | v | ڤ | پەیڤ |
| **6** | | | | z | z | ز | ڕێز |
| **7** | | | | ʒ | zh | ژ | کەژ |
| **8** | | | | ɣ | xe | غ | ساغ |
| **9** | | | | ʕ | ah | ع | بێعار |
| **10** | | Unvoiced | Stop | t | t | ت | دەسەڵات |
| **11** | | | | t͡ʃ | ch | چ | ورچ |
| **12** | | | | k | k | ک | بووک |
| **13** | | | | p | p | پ | گڵۆپ |
| **14** | | | | q | q | ق | لاق |
| **15** | | | | ʔ | eh | ء | مەسئوول |
| **16** | | | Fricative | h | h | هـ | هەناسە |
| **17** | | | | s | s | س | بسک |
| **18** | | | | ʃ | sh | ش | ڕەش |
| **19** | | | | f | f | ف | ماف |
| **20** | | | | x | x | خ | بەرخ |
| **21** | | | | ħ | he | ح | حۆل |
| **22** | | Vibrant | Flap | ɾ | r | ر | کار |
| **23** | | | Trill | r | rr | ڕ | گۆڕین |
| **24** | | Lateral | | l | l | ل | لار |
| **25** | | | | ɫ | ll | ڵ | هەوڵ |
| **26** | | Nasal | | m | m | م | کەم |
| **27** | | | | n | n | ن | کۆن |
| **28** | | Approximant | | j | y | ی | کەی |
| **29** | | | | w | w | و | چەوت |
| **30** | Vowels | Front | High | i | i | ی | نەوی |
| **31** | | Central | Low | ä | aa | ا | باران |
| **32** | | Front | Mid-low | ɛ | e | ێ | ئەوێ |
| **33** | | Back | Mid | o | o | ۆ | زۆر |
| **34** | | Central-back | Mid-high | ʊ | u | و | کورد |
| **35** | | Back | High | u | uu | وو | لووت |
| **36** | | Front | Low | a | a | ە | بەش |
| **37** | | Central-front | Mid-high | ɪ | | | دڵ |

Although the writing system of Central Kurdish is very similar to Persian and Arabic writing systems, there are several discrepancies. A comparison of letters in Kurdish, Persian and Arabic alphabets are given in Table 2. These distinct letters can be used as a feature for language identification as well (Veisi et al. 2019). As the four Arabic/Persian diacritics (ـَـُـِـّ) are not written usually, the homograph ambiguity and Kasre problems occur in Arabic and Persian (Bijankhan et al. 2011). In contrast, in the Kurdish writing system, there are specific corresponding letters for them that appear in written texts; therefore, the aforementioned types of problems do not occur in Kurdish.

Table 2: Letters of Kurdish in comparison with Persian and Arabic letters

| Language | Letters | | | | | | | | | | | | | | | | | | | | | | |
|---|---|---|---|---|---|---|---|---|---|---|---|---|---|---|---|---|---|---|---|---|---|---|---|
| **Kurdish only** | ڤ<br>/v/ | ڕ<br>/rr/ | ڵ<br>/ll/ | ە<br>/a/ | ێ<br>/e/ | ۆ<br>/o/ | | | | | | | | | | | | | | | | | |
| **Kurdish and Persian** | ژ<br>/zh/ | پ<br>/p/ | چ<br>/ch/ | گ<br>/g/ | | | | | | | | | | | | | | | | | | | |
| **Kurdish, Persian, and Arabic** | ء<br>/eh/ | ا<br>/aa/ | ب<br>/b/ | ت<br>/t/ | ج<br>/je/ | ح<br>/he/ | خ<br>/kh/ | د<br>/d/ | ر<br>/r/ | ز<br>/z/ | س<br>/s/ | ش<br>/sh/ | ع<br>/ah/ | غ<br>/xe/ | ف<br>/f/ | ق<br>/q/ | ک<br>/k/ | ل<br>/l/ | م<br>/m/ | ن<br>/n/ | و<br>/w/ | هـ<br>/h/ | ی<br>/y/ |
| **Persian and Arabic** | ث<br>/th/ | ص<br>/ṣ/ | ض<br>/ḍ/ | ذ<br>/dh/ | ظ<br>/ẓ/ | ط<br>/ṭ/ | ـَ<br>/a/ | ـِ<br>/e/ | ـُ<br>/o/ | ـّ<br>(shaddah) | | | | | | | | | | | | | |

# 3 Kurdish Speech Corpus

The primary data source for building a speech recognition system is a speech corpus, ideally a large-scale one. In this work, we introduce the first speech corpus for Central Kurdish called AsoSoft Speech Corpus. To this end, we designed a collection of sentences and recorded them by different speakers. The designed sentences include two sets, train and test, and the recording is done in two different ways, in a noise-free office using the pre-defined microphone and in social networks using cell-phone microphones. In the following sections, the design process and the data collection are introduced. Also, the detailed properties of the AsoSoft Speech Corpus for train and test sets are given.

## 3-1 Corpus Design

As our goal for creating the AsoSoft Speech Corpus is to be used in speech recognition systems, it is required that the sentence collection of the corpus covers the acoustic variations of the Kurdish language. Therefore, the corpus should desirably be a sample of real-world probabilities in terms of phoneme distribution and frequency. To this aim, we considered di-phone as the basis upon which we measure these probabilities inspired by previous works in English for TIMIT (Garofolo 1993) and in Persian for FarsDat (Bijankhan et al. 1994) corpora.

To calculate the di-phone distribution of the Central Kurdish as well as to extract words from, we leveraged the AsoSoft text corpus (Veisi et al. 2019). Using extracted words from this text corpus, a set of sentences are designed to achieve the optimal collection of sentences. The words were selected in a way that the resulting sentence collection contains all the di-phones which occur in Central Kurdish. Furthermore, the resulting sentence collection should theoretically have the same di-phone distribution as that of a given text corpus, thus ensuring that the sentence collection resembles real-world probabilities. The procedure is comprised of a repetitive cycle of selecting words, making sentences using them, and recalculation of statistics. In addition to choosing statistics-preserving words, this is accomplished through prioritizing most frequent words from the text corpus which help the di-phone statistics of the sentence collection approximate that of the text corpus. After every round of word selection and sentence making using the selected words, the statistics would slightly change due to the inter-word di-phones, i.e., di-phones made as a result of two words being located near each other. Therefore, the above procedure was done recursively until the point that we reached a reasonable number of sentences with the optimal di-phone coverage and distribution. A histogram of the relative frequency of di-phones in the corpus and the designed sentences, for the most 20 frequent di-phones, is presented in Figure 1. In the calculation of the di-phone statistics of the text corpus and the speech corpus sentence collection, we considered punctuations as a pause in speech and thus neighboring words not resulting in the creation of a di-phone.

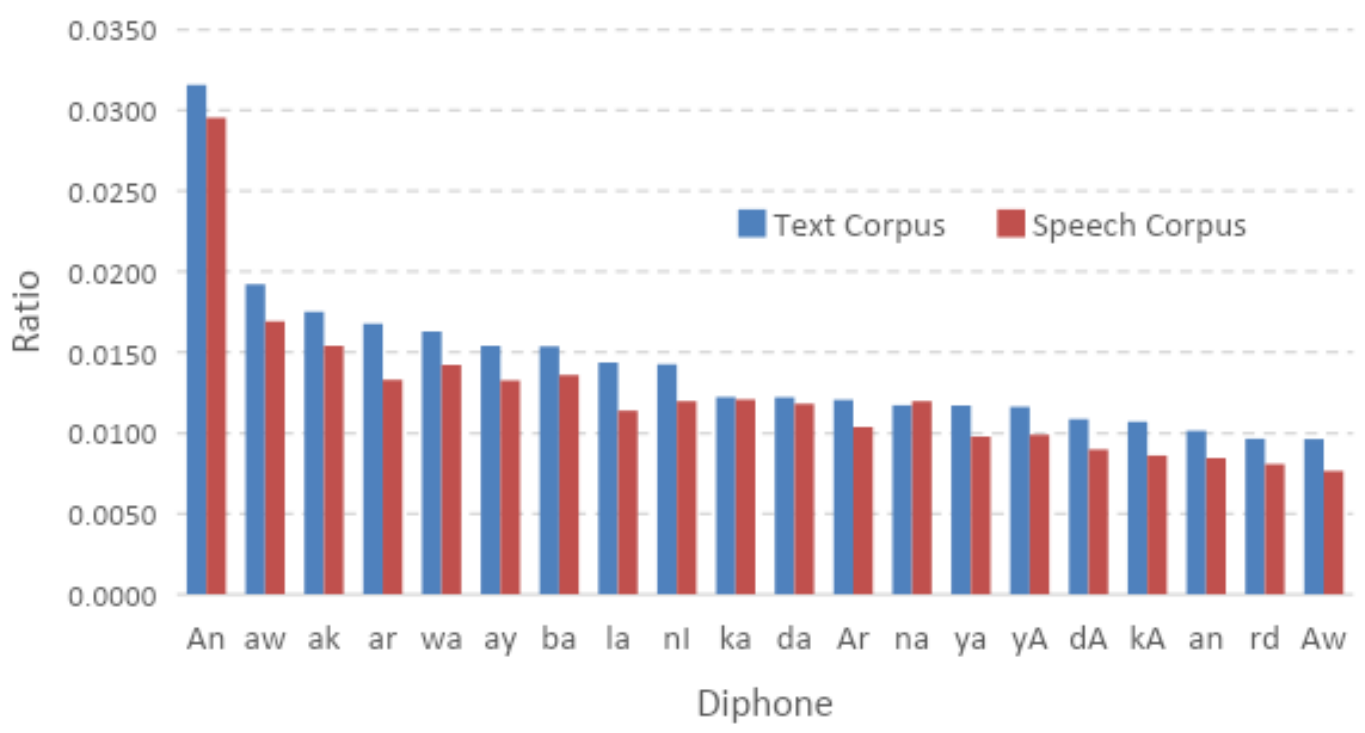

Figure 1: Comparison of di-phone ratios for 20 most frequent di-phones in AsoSoft text corpus and speech corpus sentence collection

For calculating di-phone statistics, the corpus text and collected sentences had to be transcribed into their phonetic form. We draw upon details of transcription of Central Kurdish text and challenges faced in doing so in Section 4-1. Finally, we have designed 700 sentences as the train set, two of them contain all Central Kurdish phonemes. The overall specification of the designed sentences is demonstrated in Table 3.

Table 3: The overall specification of train set sentences in AsoSoft Speech Corpus

| Title | Value |
|---|---|
| Number of sentences | 700 |
| Number of sentences contain all phonemes | 2 |
| Number of tokens | 3,950 |
| Number of tokens | 2,568 |
| Average length of each sentence | 5.6 |

Samples of the designed sentences are presented in Table 4 in which the last two rows present the sentences that contain all Central Kurdish phonemes.

Table 4: Some samples of the designed train sentences in AsoSoft Speech Corpus

| Kurdish | Phonetic Form | English |
|---|---|---|
| هیچ جۆرە دەرمانێک بۆ ئەو نەخۆشییە نەبوو | h-i-ch je-o-r-a d-a-r-m-aa-n-e-k b-o eh-a-w n-a-x-o-sh-i-y-a n-a-b-uu | There wasn't any drug for that disease |
| ملوانکە شین ئیخڕاج کرا | m-l-w-aa-n-k-a sh-i-n eh-i-x-rr-aa-je k-r-aa | The blue necklace wearer ousted. |
| بە خەباتی سەرەکیی گەل هەرێمێک پەیدا بووە | b-a x-a-b-aa-t-i s-a-r-a-k-i-y g-a-l h-a-r-e-m-e-k p-a-y-d-aa b-u-w-a | By endeavor of people a state is established |
| هەژار حەمەعەلی و خەجێ مرۆڤ چەن قفڵ و گسکی شپرزی باغیان تووڕ دا | h-a-zh-aa-r he-a-m-a-ah-a-l-I w x-a-je-e m-r-o-v ch-a-n q-f-ll uu g-s-k-i sh-p-r-z-i b-aa-xe-y-aa-n t-uu-rr d-aa | Hajar Hama Ali and Khaje Mirov threw several locks and untidy of the garden |
| غەفوور کەمعورزە و ڕێواس جلپێهات حەڤدە قووچی برژیاگیش ئەخۆن | xe-a-f-uu-r k-a-m-ah-u-r-z-a w rr-e-w-aa-s je-l-p-e-h-aa-t he-a-v-d-a q-uu-ch-i b-r-zh-y-aa-g-i-sh eh-a-x-o-n | The dummy Ghafour and beauty Rewas eat seventeen baked rams |

Although a subset of the mentioned sentences can be used as the test set in a speech recognition system, however, a competitive speech recognition system must have a good performance in different domains, especially for the sentences other than the training ones. Therefore, we have also designed a sentence collection for the evaluations of ASR systems. To build the test sentences, we extracted 100 sentences from a set of documents in 11 different domains, i.e., religious, sport, politics, economics, science and technology, social, novels, poet, formal letters, conversation, and general. These sentences are extracted from different online sources and then they are refined. The topics of the selected sentences and the number of sentences for each topic are given in Table 5.

Table 5: Distribution of the designed test set sentences over different domains

| **Topics** | **Number of Sentences** |
|---|---|
| General | 10 |
| Religious | 10 |
| Sport | 10 |
| Politics | 10 |
| Economics | 10 |
| Social | 10 |
| Novel | 10 |
| Letter | 10 |
| Conversation | 10 |
| Scientific/Technology | 5 |
| Poet | 5 |
| **Total** | **100** |

## 3-2 Data Collection

In this section, we describe the process of speech data collection and present the specification of the recorded dataset. The designed speech sentences described in the previous section are recorded in two different environments resulting in two different subsets for the corpus: 1) Controlled office environment with noise-free microphone, called AsoSoft Speech-Office; and 2) Various acoustic environments with cell-phone microphones using crowdsourcing in Telegram social network, called AsoSoft Speech-Crowdsourcing. In both environments, the train and test data are recorded. After the recording, all samples of the data are manually processed to discard noisy and corrupted samples.

The train speech corpus is 43.68 hours of duration. A general summary of the training dataset is given in Table 6. The detailed description of these subsets is given in the following.

Table 6: AsoSoft Speech Corpus (train set)

| | **No. of Speakers** | **No. of Utterances** | **Duration (Hours)** |
|---|---|---|---|
| Speech-Office | 60 | 31,075 | 31.52 |
| Speech-Crowdsourcing | 516 | 11,519 | 12.16 |
| **Total** | **576** | **42,594** | **43.68** |

### 3-2-1 AsoSoft Speech-Office

In the AsoSoft Speech-Office, the recording conditions, including the speaker information and hardware setup are known. Therefore, a special format is used in the naming the speech corpus files:

- the first character shows the gender of the speaker, F is for female and M for male;
- the second character is a 0 or 1 showing the recording device, where 0 denotes that the file is recorded by laptop and 1 shows that it is recorded by PC;
- the third character shows the type of microphone, where 0 means a USB microphone (we have used Andrea NC-181VM USB) and 1 indicates that the microphone is connected with jack port (VXI UC Proset headset is used);
- the next three characters show the sentence ID that is a decimal number from 000 to 699;
- and the last three characters show the speaker ID (Table 7).

The speech files are recorded in wave format, 22050 Hz, 16 bits, mono. For each .wav file, a .txt file and a .phn file are also produced with the same name containing the transcription and pronunciation of the recorded sentences, respectively.

Table 7: File format of AsoSoft Speech-Office

| Gender | Lap/PC | USB /JACK | Sentence ID | Speaker ID |
|---|---|---|---|---|
| F/M | 0/1 | 0/1 | 000-699 | 000-999 |

The major part of the recorded training data in the corpus belongs to AsoSoft Speech-Office which is 31.5 hours. In this subset, each speaker has uttered a subset of (or all) the designed sentences. Those two sentences that include all 37 Central Kurdish phonemes are uttered by all speakers. A summary of the AsoSoft Speech-Office corpus specifications is given in Table 8. Among 60 speakers, 39 of them have uttered all 700 sentences and the average number of utterances for each speaker is about 518 sentences.

The diversity of speech data is crucial to have a robust ASR system. Therefore, we tried to record the speech data from both genders, speakers with different education levels, different ages, and speakers from different dialects and sub-dialects of the Kurdish language. The detailed information about gender, education, age, and dialects of speakers is presented in Figure 2. As it is shown in this figure, the AsoSoft Speech-Office subset is nearly balanced for genders; most speakers are middle-aged; and more than half of them are the speakers of Central Kurdish.

Table 8: Specification of AsoSoft Speech-Office (train)

| Title | Value |
|---|---|
| **Dataset Name** | AsoSoft Speech-Office Train V1 |
| **Recording** | Andrea NC-181VM USB |
| **Microphone** | VXI UC Proset |
| **Duration (Hours)** | 31.5 |
| **Num. of Speakers** | 60 |
| **Num. of Utterances** | 31,075 |
| **Average Utterance Length (Seconds)** | 3.65 |
| **Average Utterance Per Speaker** | 517.9 |
| **Frequency** | 22.05 kHz |
| **Sampling Resolution** | 16 Bit, Mono |
| **Format** | MS Wav |

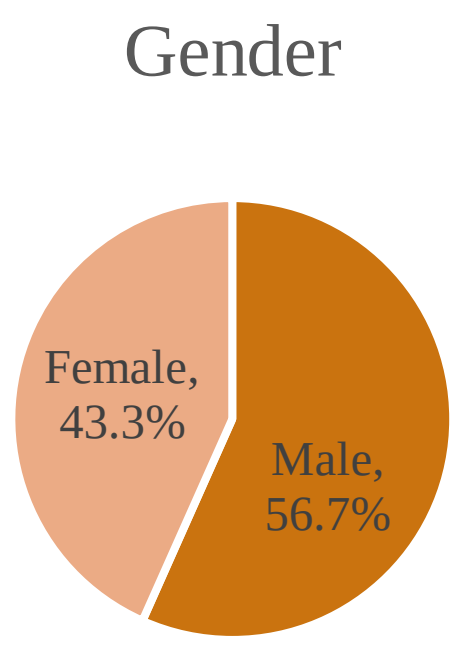

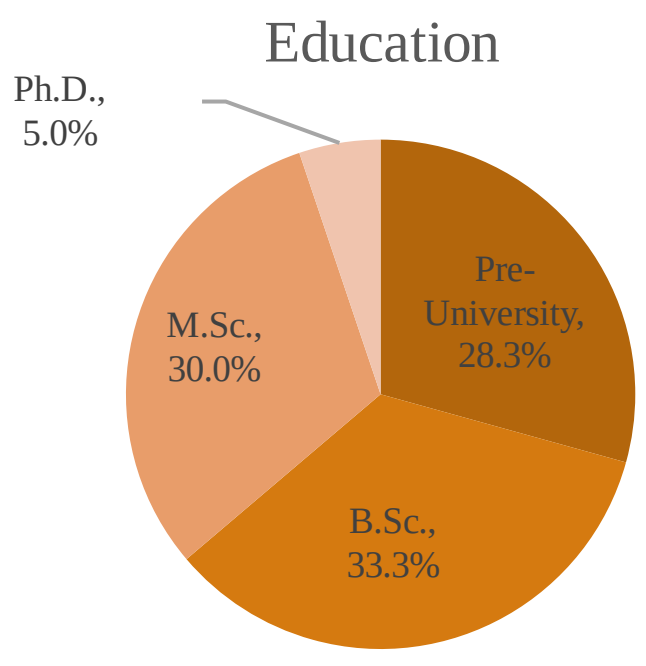

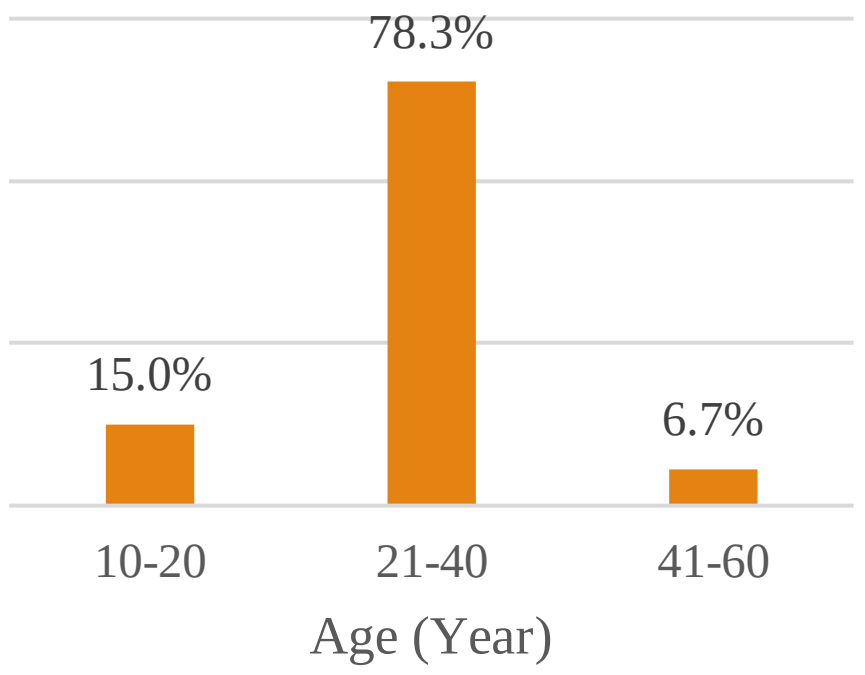

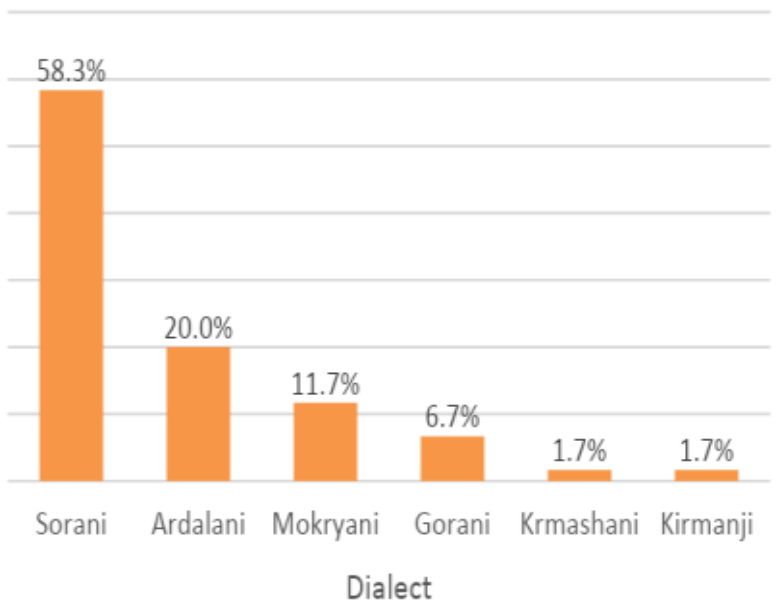

Figure 2: Information of speakers in AsoSoft Speech-Office (train)

### 3-2-1 AsoSoft Speech-Crowdsourcing

The second part of the corpus is collected using a designed crowdsourcing bot in Telegram social network. In the time of creating the corpus, Telegram has been the most popular social network in Iran and is widely available on Kurd's cell-phones as well. The designed bot shows a sentence randomly from the 700 designed sentences and also plays its related voice, then asks the Telegram user to utter it. The user can repeat a recorded sentence or go to the next sentence. Also, we have asked for the user's information, but to simplify the recording process, they are not mandatory. Therefore, the information for age, education, and dialect of the speakers is not available for all of them. However, we have labeled the gender of the speakers by manual checking the samples in which 77.9% of the utterances belong to males and 22.1% of them are uttered by females. The specifications of this subset are presented in Table 9 in which the duration is 12.16 hours uttered by 516 speakers. The audio recording format of this subset is the same as the AsoSoft Speech-Office and the txt and phn files are also available for them.

Table 9: Specification of AsoSoft Speech-Crowdsourcing (train)

| Title | Value |
|---|---|
| **Dataset Name** | AsoSoft Speech-Crowdsourcing Train V1 |
| **Recording Device** | Microphone of Cell-phones |
| **Duration (Hours)** | 12.16 |
| **Number of Speakers** | 516 |
| **Number of Utterances** | 11,519 |
| **Average Utterance Length (Seconds)** | 3.65 |
| **Average Utterance Per Speaker** | 22.3 |
| **Frequency** | 22.05 kHz |
| **Sampling Resolution** | 16 Bit, Mono |
| **Format** | MS Wav |

According to the fact that there is not any control over the recording condition in this subset, it is processed manually and corrupted samples are removed. This manual cleaning reduced the number of speakers from 558 to 516 and also decreased the number of utterances from 12,917 into 11,519 which results in discarding about 2 hours of the recorded speech (from 14.06 to 12.16 hours).

### 3-2-3 AsoSoft Speech Test Set

As mentioned, we have designed a set of sentences for the test set and they are recorded in both Office and Crowdsourcing environments. In the Office environment, 8 speakers have uttered all 100 sentences of Table 5, resulting in 1.25 hours of speech. The recording conditions for this test set is the same as the recording condition of the AsoSoft Speech-Office train set. The second subset of the test samples, i.e., Crowdsourcing, is collected in the same manner as the AsoSoft Speech-Crowdsourcing train subset. The designed 100 sentences are randomly demonstrated to Telegram users and they are uttered them. In a similar manner as the Crowdsourcing train set, here the number of speakers and acoustic variations are high but the average utterance per speaker is low (about 4 sentences for each speaker). For this subset, after removing the distorted samples, 135 speakers have uttered 558 sentences, resulting in more than 50 minutes of speech. It should be noted that the distribution of the utterances over 11 topics of the test sentences for this subset is similar to that given in Table 5.

Table 10 summarizes the specifications and statistics of our collected test data in two subsets.

Table 10: Overall specification of AsoSoft Speech-Test set

| Title | Office | Crowdsourcing | Total |
|---|---|---|---|
| **Dataset Name** | AsoSoft Speech-Test | | |
| **Recording Microphone** | Andrea NC-181VM USB/VXI UC Proset | Cell-phones | - |
| **Duration (Hours)** | 1.25 (01:15':43'') | 0.84 (00:50':17'') | 2.09 (02:06':00'') |
| **Number of Speakers** | 8 | 135 | 143 |
| **Number of Utterances** | 800 | 558 | 1,350 |
| **Average Utterance Length (Seconds)** | 5.7 | 5.41 | 5.55 |
| **Average Utterance Per Speaker** | 100 | 4.1 | 9.4 |
| **Frequency** | 22.05 kHz | | |
| **Sampling Resolution** | 16 Bit, Mono | | |
| **Format** | MS Wav | | |

## 4 Kurdish pronunciation Lexicon

The pronunciation lexicon as a key data resource plays an important role in speech recognition systems. It imposes to the system that only the orthographically correct forms are used. Lexicon is generally extracted from a very large collection of text, i.e., a text corpus. A lexicon contains unique entries each of which corresponds to a token in the text corpus. Therefore, tokenization and normalization are major steps in extracting lexicon from the text corpus. Additionally, in a lexicon used for the speech recognition system, the entries must be phonetically transcribed. In this section, we describe the steps taken for extracting our lexicon and the challenges faced alongside their solutions.

To extract the lexicon entries, we opted for AsoSoft Text Corpus which is the first large-scale Central Kurdish text corpus (Veisi et al. 2019). This text corpus contains 188 million tokens composed of 4.66 million unique tokens. We primarily have selected 100K of the most frequent tokens as the lexicon which reduced to 60K after handly corrections and script standardizations. To this aim, the latest version of AsoSoft corpus is adopted which was preprocessed, normalized and different challenges facing Central Kurdish script standardization had been coped with (Mahmudi et al. 2019; Mohammadamini et al. 2019). In the following, we draw upon steps taken for automatic phonetically transcribing the lexicon entries. This algorithm has also been exploited to extract di-phone statistics for developing the speech corpus sentences, as described in Section 3-1. Also, after reviewing the literature on Central Kurdish script standardization, we developed standardization rules for Central Kurdish script, as can be seen in Section 4.2. Three experts in Central Kurdish examined the correctness of the lexicon pronunciations and scripts and applied the necessary corrections.

### 4-1 Phonetic Transcription

A significant aspect of a lexicon in an ASR system is the phonetic transcription of entries. Although the Kurdish orthography tend to be phonemic and almost every letter corresponds to one phoneme, there are some challenging exceptions. Two letters "ی" and "و" have more than one form of pronunciation: "ی" can be pronounced both as /j/ (palatal approximant) and as /i/ (a vowel); and "و" could be pronounced both as /w/ (bilabial approximant) and as /ʊ/ (a short vowel). Additionally, the letter "و" can also be written in a repeated form as the digraph "وو" and is pronounced as /u/ (a long vowel). These variations result in different transcription forms as well as different di-phone formations.

Considering these uncertainties and the lack of pronunciation dictionaries for low-resourced language of Central Kurdish, for converting the lexicon entries into their equivalent pronunciations we adopted the rule-based grapheme-to-phoneme (G2P) conversion method presented by Mahmudi and Veisi (2020). Using the rules of phonology and syllabification of Central Kurdish, this G2P convertor generates all possible pronunciations of each word. Then, each candidate is evaluated if violates the phonological constraints of the language. Each violation increases the penalty value of that candidate. Eventually, a candidate with the lowest amount of penalty is returned as the most probable well-formed pronunciation of the written word. This algorithm is 100% accurate in the conversion of the letters "ی" and "و" as reported (Mahmudi and Veisi 2020).

**4-2 Script Standardization**

Despite several attempts, the Kurdish orthography has not been fully standardized yet. As a result, there are different spellings for many Central Kurdish words. Therefore, it is necessary to determine unique written forms as the output of the speech recognition engine. In Table 11, we have categorized the instances of verb lexemes that could be written in different forms in Central Kurdish. Table 12 gives similar examples for nouns and adjectives. In these tables, we have proposed the standard form for each case. To be confident in selecting the standard form, various references and related literatures on the Kurdish script standardization are surveyed (Baban 2008; Khoshnaw 2013; Mohammadamini et al. 2019; The Kurdish Academy 2010). There are some cases that there is no consensus among the researchers on selecting the standard form in which we have selected the standard form based on several clues such as the majority voting on the suggestions by various references(Baban 2008; Khoshnaw 2013; The Kurdish Academy 2010), the frequency of the forms in the text corpus (Veisi et al. 2019), prioritizing the written language instead of informal spoken language and also the language intuition of the linguists. After determining the standard form for each instance, we have manually corrected lexicon entries accordingly.

Table 11: Script Standardization for Verbs in Central Kurdish

| Verb Tense | Morpheme | Allomorph | Standardized form | Other forms |
|---|---|---|---|---|
| Imperative | Suffix ‘ەوە’ after a vowel | ڕەوە/وە | بیخەڕەوە<br>بینێرەڕەوە | بیخەوە<br>بینێرەوە |
| Present/ Imperative | Subjunctive/ Imperative prefix | ب=0 | دابکەون | داکەون |
| Present | Progressive affix | دە/ئە | دەکەوم | ئەکەوم |
| Present | Clitic ‘ش’ after vowel | ش/یش | دەشکەوم | دەیشکەوم |
| Present | Second-person singular pronoun | یت/ی | دەکەویت | دەکەوی |
| Present | Third-person singular pronoun | ێت/ێ<br>ات/ا (stem ends with /ا/ or /ۆ/)<br>ت/0 (stem ends with /ێ/) | دەکەوێت<br>دەخوات<br>دەڵێت | دەکەوێ<br>دەخوا<br>دەڵێ |
| Present | Passive suffix in present tense | درێ/رێ (after /ڕ/)<br>درێ/رێ (after /ڕ/)<br>درێ/رێ (after /ێن/) | بکرڕدرێت<br>بگرڕدرێت<br>دەقرتێندرێت | بکرڕرێت<br>بگرڕرێت<br>دەقرتێنرێت |
| Present | The blending of progressive affix with stems of verbs which start with ‘هێ’ | دەهێ/دێ | دەهێنێت<br>دەهێڵێت | دێنێت<br>دێڵێت |
| Present | The blending of subjunctive/imperative affix with stems of verbs which start with ‘هێ’ | بهێ/بێ | بهێنم<br>بهێڵم | بێنم<br>بێڵم |
| Past | Progressive affix | دە=ئە | دەکەوتم | ئەکەوتم |
| Past | Clitic ‘ش’ after vowel | ش/یش | دەشکەوتم | دەیشکەوتم |
| Past | Suffix ‘ەوە’ in the present perfect tense | ووەتەوە=ۆتەوە | کردووەتەوە | کردۆتەوە |
| Past | The past perfect tense of the verb ‘بوون’ | | بووبوو | ببوو |
| Past | Passive suffix in past tense | درا/را (after /ێن/)<br>درا/را/ا (after /ڕ/)<br>ردرا/را (present stem ends with /ە/) | قرتێندرا<br>گۆڕا<br><br>بردرا | قرتێنرا<br>گۆڕرا ،<br>گۆڕدرا<br>برا |
| Past | Suffix ‘ەوە’ after verbs which end in a vowel | یەوە/وە | دایەوە | داوە |
| Past | Third-person singular in the present perfect tense (when subject pronoun at the end of the verb isn’t replaced) | وویەتی/وویە | ناردوویەتی | ناردوویە |

Table 12: Script Standardization for Nouns and Adjectives in Central Kurdish

| Morpheme | Allomorphs | Standardized Form | Other Forms |
|---|---|---|---|
| Suffix ‘دا’ | دا=ا<br>دا=یا | ماڵدا<br>برادا<br>تێدا | ماڵا<br>برایا<br>تیا |
| Indefinite suffix | ێک=ێ<br>یەک=یێک/یێ<br>یەک=ەک/ێک | ماڵێک<br>برایەک<br>خانوویەک | ماڵێ<br>برایێک، برایێ<br>خانووەک، خانوویێک |
| Definite suffix in words which end in a vowel | کە=یەکە | براکە | برایەکە |
| Suffix ‘ەوە’ in words which end in a vowel | یەوە=وە | لەو کێشەیەوە | لەو کێشەوە |
| Morphemes which end in ‘ست’ | ست=س | دەست | دەس |
| Suffix ‘ی’ in words which end in ‘ی’ | ی=0 | ئابووریی | ئابووری |
| Suffix ‘انە’ in words which end in ‘ی’ | یانە=انە | ئابوورییانە | ئابووریانە |
| Deletion in compound words | ئا=ا<br>هە=ە<br>وو=و | ڕۆژئاوا<br>ڕۆژهەڵات<br>هاووڵاتی | ڕۆژاوا<br>ڕۆژەڵات<br>هاوڵاتی |
| ‘و’ at the beginning of words | و=وو | وشە | ووشە |
| Suffix ‘ێتی’ | ێتی/یەتی/ەتی | یەکێتی | یەکیەتی، یەکەتی |
| Location suffix | گە/گا | ڕێگە/ڕێگا | ڕێگە/ڕێگا |

In Central Kurdish, similar to some other languages, a great proportion of foreign loan-words and proper names have several variations. For these words, the most frequent one in the text corpus (Veisi et al. 2019) is selected as the main entry and the pronunciations of other forms are added to the pronunciation lexicon as well. Some examples of these words and their frequencies in the Asosoft text corpus is presented in Table 13.

Table 13: Examples of variations in Kurdish loan-words and proper names along with their frequencies

a) Istanbul

| Variation | Frequency |
|---|---|
| ئەستەمبووڵ | 1,486 |
| ئەستەنبوڵ | 760 |
| ئەستەمبوڵ | 525 |
| ئیستانبوڵ | 272 |
| ئەستەموڵ | 267 |
| ئیستانبول | 252 |
| ئیستەنبوڵ | 176 |
| ئیستانبووڵ | 148 |
| ئیستەنبول | 74 |
| ئیستانبوول | 40 |
| ئەستەمبوول | 31 |
| ئەستەمول | 19 |
| ئەستامبوڵ | 11 |
| ئێستانبول | 7 |
| ئێستانبوڵ | 6 |
| ئەستامبول | 4 |
| ئیستەمبول | 4 |
| ئیستەمبوڵ | 2 |
| ئێستانبووڵ | 2 |
| ئیستەمبوول | 2 |
| ئیستەمبووڵ | 1 |
| ئەستامبووڵ | 1 |

b) Hydrogen

| Variation | Frequency |
|---|---|
| هایدرۆجین | 225 |
| هایدرۆژین | 18 |
| هایدروجین | 9 |
| هیدرۆژن | 6 |
| هیدرۆجین | 3 |
| هیدرۆژین | 1 |

c) Culture

| Variation | Frequency |
|---|---|
| کولتوور | 3,839 |
| کەلتوور | 2,455 |
| کلتور | 1,053 |
| کولتور | 918 |
| کلتوور | 895 |
| کەلتور | 866 |

d) Mohammad

| Variation | Frequency |
|---|---|
| محەمەد | 64,521 |
| موحەممەد | 32,014 |
| موحەمەد | 4,009 |
| محەممەد | 111 |
| محەمە | 99 |
| مۆحەمەد | 10 |
| مۆحەمەد | 6 |

# 5 Jira Kurdish Speech Recognition

As another contribution of this work, for the first time, we have developed a speech recognition system for Central Kurdish. To implement the system, we have used Kaldi toolkit (Povey et al. 2011) using three data sources: 1) AsoSoft text corpus (Veisi et al. 2019) for language modeling, 2) AsoSoft speech corpus introduced in Section 3 to create acoustic models and also evaluate the system, and 3) the pronunciation lexicon presented in Section 4. Kaldi is an open-source speech recognition toolkit that can be used to produce different language models and acoustic models. In Section 5-1 and Section 5-2, training the acoustic model and language model in Kaldi framework are described, respectively. Then, the evaluation results of the system are given in Section 5-3.

## 5-1 Acoustic Modelling

To create acoustic models, for each frame (20ms and 50% overlap) of an utterance, 13 Mel-Frequency Cepstral Coefficients (MFCCs) are extracted, then the MFCCs are normalized with Cepstral Mean-Variance Normalization (CMVN) method. To do the normalization, each frame is normalized using the statistics of 4 previous frames and 4 next frames. Various configurations of the system have been evaluated in some of them the dynamic features. i.e., delta and delta-delta are also appended to the feature vector. Also, Linear Discriminative Analysis (LDA), Maximum Likelihood Linear Transform (MLLT) and Speaker Adaptive Training (SAT) are applied on MFCC features to transform the speech frames in some other configurations. The following configurations are used in this paper to construct the acoustic model:

- Mono: This denotes the HMM-based mono-phone modeling using Gaussian Mixture Models (GMM) and MFCC features. The number of models for this case is 37 (we have not modeled the short vowel /i/ and also, we have modeled a silence model).

- Tri1: It defines HMM-GMM-based context-dependent tri-phone modeling using MFCC, delta, and delta-delta features, resulting in 39-dimensional feature vectors. The number of senones, in this case, is 2,500.
- Tri2: In this experiment, HMM-GMM-based tri-phone modeling (by 2,500 senones) using MFCC features transformed by LDA and MLLT into a 40-dim vector is performed.
- Tri3: The configuration of this case is the same as the Tri2 but SAT is also performed.
- SGMM: This model denotes the Subspace Gaussian Mixture Model (SGMM) that drives the global shared model with low dimensional state vector. In our experiments, the LDA transform is applied to the MFCC features and the number of intermediate Gaussians is 2,000.
- DNN: In this case, the acoustic modeling is performed using Deep Neural Networks (DNN) and HMM. The network depth is 6 and in each of them, 1,024 neurons are used.

The evaluation results for each of the mentioned acoustic modeling methods are presented in Section 5-3.

### 5-2 Language Modelling

The language model is another important data source for recognizing final words in an ASR system that constrains search space in the decoder. We have used a statistical n-gram model in Jira. To extract the language model in our system, the AsoSoft text corpus (Veisi et al. 2019) is used. This text corpus is mainly collected from books, magazines, newspapers and online websites, and contains 188 million tokens. Text normalization is a critical step to improve the quality of language models and in the case of the Kurdish language, this step is more critical since different encoding systems and various orthographic rules are used by Kurdish writers and publishers. The details of normalization done on the text corpus are discussed in (Mahmudi and Veisi 2020; Veisi et al. 2019). Furthermore, a correction table which is produced during the lexicon preparation and script standardization (in Section 4-2), is applied to the text corpus to enrich the normalization.

To construct the statistical language model of the Jira system, the Kaldi toolkit is used. As mentioned in Section 4, the lexicon size is 60K and the trigram statistical language model with back-off smoothing is extracted.

### 5-3 Evaluations and Results

In this section, the results of our experiments in setting up Jira Kurdish ASR are presented. As described in Section 3, our speech corpus for both train and test includes two subsets, Office and Crowdsourcing. In the following, we have presented the recognition performance for training the system using these two subsets separately, also their combination as a single set (denoted as Office+Crowdsourcing). For each of these cases, the evaluations are done on both test sets, i.e., Office-Test and Crowdsourcing-Test. In the evaluations, two criteria including WER (word error rate) and PER (phoneme error Rate) are reported.

In all experiments, the results of different speech recognition acoustic models including Mono, Tri1, Tri2, Tri3, SGMM, and DNN are given. Table 14 presents the results of Jira on Office-Test for three different training conditions. The reported rates are the average of the evaluation criteria over all samples of the test set. As it is observed, the lowest PER for this test set is achieved by SGMM and DNN (PER is 10%) for Office and Office+Crowdsourcing training sets, respectively. Also, the lowest WER is resulted by the SGMM method for the Office train set (i.e., 13.9%). The average performance for Office-Test at the last row also confirms the superiority of Office training condition for WER. The higher performance of the system for the Office test set using the Office train set is not supervising due to the matching acoustic condition between the train and test. The performance of the system for the Crowdsourcing training condition is lower than the two other cases due to the acoustic mismatch. Besides, DNN has achieved better performance for PER on Office+Crowdsourcing training condition which is probably due to the larger training samples. Generally, the low WER of the system is probably due to the mismatch in the language model between the train text corpus (Veisi et al. 2019) which is mainly from web sites and the test set which include various genres.

Table 14: PER and WER of Jira ASR system on Office-Test set

| **Trainset⇨** | **Office** | | **Crowdsourcing** | | **Office+Crowdsourcing** | |
|---|---|---|---|---|---|---|
| **Acoustic Model⇩** | **PER** | **WER** | **PER** | **WER** | **PER** | **WER** |
| Mono | 33.4 | 28.0 | 37.9 | 30.3 | 32.3 | 27.7 |
| Tri1 | 19.8 | 18.1 | 24.9 | 23.0 | 19.3 | 18.3 |
| Tri2 | 16.1 | 16.5 | 22.7 | 22.5 | 17.0 | 17.1 |
| Tri3 | 16.0 | 15.0 | 18.2 | 22.1 | 13.2 | 16.3 |
| SGMM | 10.0 | 13.9 | 12.5 | 15.9 | 10.6 | 15.7 |
| DNN | 12.0 | 16.7 | 12.2 | 19.3 | 10.0 | 17.3 |
| **Average** | **17.9** | **18.0** | **21.4** | **22.2** | **17.1** | **18.7** |

A similar evaluation is also performed on Crowdsourcing-Test which the results are shown in Table 15. According to what is represented in this table, the SGMM model trained by Crowdsourcing training set outperforms the other cases in terms of WER (which is 14.4%) and PER (which is 11.9%). This superiority originates from the matched acoustic condition for collecting the data using crowdsourcing in both train and test sets. Due to the high acoustic variations of this test set, the performance of the system for Office training case is noticeably lower than the Crowdsourcing training condition.

Table 15: PER & WER of Jira ASR system on Crowdsourcing-Test set

| **Trainset⇨** | **Office** | | **Crowdsourcing** | | **Office+Crowdsourcing** | |
|---|---|---|---|---|---|---|
| **Acoustic Model⇩** | **PER** | **WER** | **PER** | **WER** | **PER** | **WER** |
| Mono | 43.2 | 33.3 | 29.2 | 28.3 | 36.4 | 30.3 |
| Tri1 | 22.2 | 22.3 | 16.4 | 18.1 | 21.8 | 25.0 |
| Tri2 | 21.3 | 20.7 | 17.3 | 17.9 | 20.4 | 23.4 |
| Tri3 | 21.1 | 20.1 | 16.1 | 17.3 | 19.1 | 23.9 |
| SGMM | 12.1 | 16.3 | 11.9 | 14.4 | 14.2 | 14.9 |
| DNN | 13.2 | 20.2 | 12.3 | 14.6 | 13.4 | 19.3 |
| **Average** | **22.2** | **22.2** | **17.2** | **18.4** | **20.9** | **22.8** |

As mentioned in Section 3-1, our test set is designed to cover various text genres (i.e., 11 topics). In Figure 3, the WER values of the system are demonstrated for 11 topics of the test set. The results of this figure are for the SGMM model trained and evaluated by the Office subset. According to the results given in Table 14 and Table 15, the lowest WER is achieved by the SGMM model trained and evaluated by Office data. As shown in this figure, the performance of the system for the *general* topic is outstandingly better than the others, probably due to the matching in the language model for this genre. On the other hand, the WER is remarkably high for the *poet* due to the LM mismatch as well as rhythmic style in reading the sentences. The other cases such as *scientific* and *religious* also suffer from enough training samples in the text corpus for train the LM.

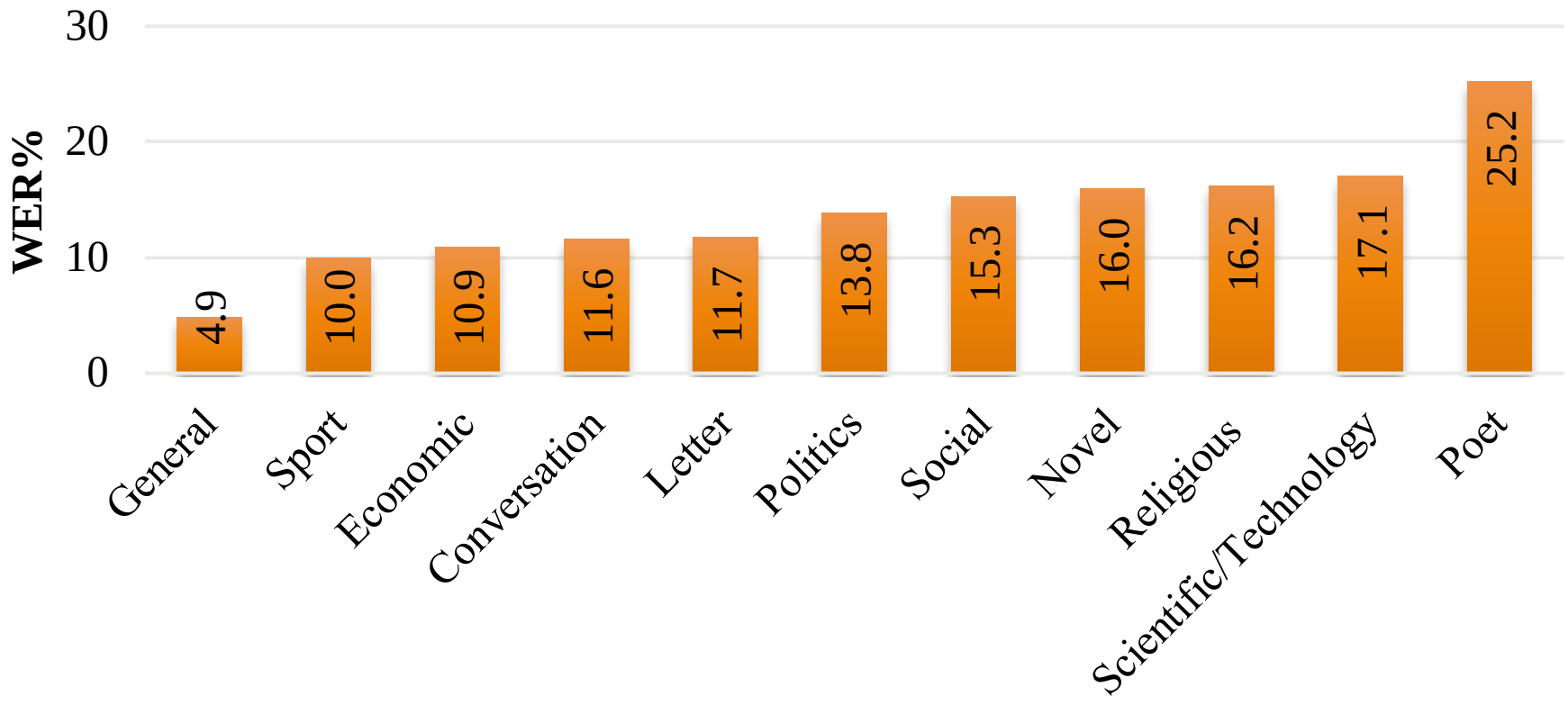

Figure 3: WER of Jira ASR system for different topics (SGMM model trained and evaluated by Office data)

## 6 Summary and Conclusions

In this paper, we introduced our experiments in designing and implementing the first Central Kurdish (Sorani) speech recognition system and its related requirements including speech corpus and pronunciation lexicon. In this research, we designed the first speech corpus for the Kurdish language based on the distribution of di-phones in the Kurdish language. In this corpus, 700 sentences for the train set and 100 sentences (covering 11 different topics) for the test set were designed and speech samples were collected in two manners: Office and Crowdsourcing. The final train set includes more than 42,000 utterances spoken by 576 speakers resulted in a corpus including more than 43 hours of speech samples. For the test set, more than 2 hours of speech were collected by 143 speakers. The Jira recognition system is designed to be used in different domains; therefore, the evaluation set includes various domains. Also, the first pronunciation lexicon extracted and standardized orthographically to be used in Kurdish speech recognition systems. The final version of the lexicon includes 60K entries in which an automatic grapheme-to-phoneme algorithm is used for generating the pronunciations. Then, the Kaldi ASR toolkit is used as the recognition engine to set up the system. Various acoustic models and the trigram statistical language model are used. The evaluations were performed on both Office and Crowdsourcing test sets using three training sets: Office, Crowdsourcing and the combination of these two sets. The experiments show that the performance of the SGMM model outperforms other acoustic models in general. As it was expected, the performance of the system for the matched acoustic conditions was better than the other cases, in which the system resulted in 13.9% of WER for the Office and 14.4% of WER for the Crowdsourcing. The performance of the system for some topics in the test set was better (for the general topic, 4.9% of WER) than some others (for the poet the WER was 25.2%). In the cases that the number of training samples was increased, i.e., the combination of Office and Crowdsourcing subsets, the DNN model resulted in better PER (10%). Also, Crowdsourcing samples include high acoustic variations (speakers, recording microphones and environment noises) which could result in higher robustness.

The authors of this paper hope that this research opens the doors to do research and implementing more practical speech recognition applications in the Kurdish language. In the future, we have planned to improve the quality of language resources for the Kurdish language and move toward applying other state-of-the-art methods such as deep learning techniques to improve the performance of the Jira recognition system.

## References

Abdul, Z. K. (2019). Kurdish speaker identification based on one dimensional convolutional neural network. *Computational Methods for Differential Equations*, *7*(4 (Special Issue)), 566–572.

Abdulrahman, R. O., Hassani, H., & Ahmadi, S. (2019). Developing a Fine-Grained Corpus for a Less-resourced Language: the case of Kurdish, 4–7. http://arxiv.org/abs/1909.11467

Al-Talabani, A., Abdul, Z., & Ameen, A. (2017). Kurdish Dialects and Neighbor Languages Automatic Recognition. *ARO-The Scientific Journal of Koya University*, *5*(1), 20–23. https://doi.org/10.14500/aro.10167

Amodei, D., Ananthanarayanan, S., Anubhai, R., Bai, J., Battenberg, E., Case, C., et al. (2016). Deep speech 2: End-to-end speech recognition in english and mandarin. In *International conference on machine learning* (pp. 173–182). PMLR.

Baban, S. (2008). *Zimanî Nûsînî Rêzimandar*. Erbil, Iraq: Aras Publisher.

Barkhoda, W., ZahirAzami, B., Bahrampour, A., & Shahryari, O.-K. (2009). A comparison between allophone, syllable, and diphone based TTS systems for Kurdish language. In *2009 IEEE International Symposium on Signal Processing and Information Technology (ISSPIT)* (pp. 557–562). IEEE.

Baum, L. E. (1972). An inequality and associated maximization technique in statistical estimation for probabilistic functions of Markov processes. *Inequalities*, *3*(1), 1–8.

Bijankhan, M., Sheikhzadegan, J., & Roohani, M. R. (1994). FARSDAT-The speech database of Farsi spoken language. In *Proceedings of International Conference on Speech Sciences and Technology* (p. Vol. 2, 826-829). PROCCEDINGS AUSTRALIAN CONFERENCE ON SPEECH SCIENCE AND TECHNOLOGY.

Bijankhan, M., Sheykhzadegan, J., Bahrani, M., & Ghayoomi, M. (2011). Lessons from building a Persian written corpus: Peykare. *Language resources and evaluation*, *45*(2), 143–164.

Department of IT of Kurdistan Regional Government. (2014). Unicode Standard for Kurdish Language. http://unicode.ekrg.org/ku_unicodes.html. Accessed 24 January 2021

Furui, S. (2005). 50 years of progress in speech and speaker recognition research. *ECTI Transactions on Computer and Information Technology (ECTI-CIT)*, *1*(2), 64–74.

Garofolo, J. S. (1993). Timit acoustic phonetic continuous speech corpus. *Linguistic Data Consortium, 1993*.

Graves, A., & Jaitly, N. (2014). Towards end-to-end speech recognition with recurrent neural networks. In *International conference on machine learning* (pp. 1764–1772). PMLR.

Haig, G. (2002). The corpus of contemporary kurdish newspaper texts (CCKNT): a pilot project in corpus linguistics for Kurdish. *Kurdische Studien*, *1*(2), 148–155.

He, Y., Sainath, T. N., Prabhavalkar, R., McGraw, I., Alvarez, R., Zhao, D., et al. (2019). Streaming end-to-end speech recognition for mobile devices. In *ICASSP 2019-2019 IEEE International Conference on Acoustics, Speech and Signal Processing (ICASSP)* (pp. 6381–6385). IEEE.

Hinton, G., Deng, L., Yu, D., Dahl, G. E., Mohamed, A., Jaitly, N., et al. (2012). Deep neural networks for acoustic modeling in speech recognition: The shared views of four research groups. *IEEE Signal processing magazine*, *29*(6), 82–97.

Hosseini, H., Veisi, H., & MohammadAmini, M. (2015). KSLexicon: Kurdish-Sorani Generative Lexicon. In *The First National Conference on Corpus-based Linguistics* (pp. 33–50).

Huang, X., Acero, A., Hon, H.-W., & Reddy, R. (2001). *Spoken language processing: A guide to theory, algorithm, and system development*. Prentice hall PTR.

Ilkhanizade, M. (2006). *Āmuzeš-e Xanden va Neveštan-e Kordi be Suret-e Elmi*. Kurdistan Publication.

Jelinek, F. (1976). Continuous speech recognition by statistical methods. *Proceedings of the IEEE*,

*64*(4), 532–556.

Khoshnaw, N. A. (2013). *Zimanî Standardî Kurdî*. Erbil, Iraq: Hêvî.

Levinson, S. E., Rabiner, L. R., & Sondhi, M. M. (1983). An introduction to the application of the theory of probabilistic functions of a Markov process to automatic speech recognition. *Bell System Technical Journal*, *62*(4), 1035–1074.

Mahmudi, A., & Veisi, H. (2020). *Automated Grapheme-to-Phoneme Conversion for Central Kurdish based on Optimality Theory*. *Under Review*.

Mahmudi, A., Veisi, H., MohammadAmini, M., & Hosseini, H. (2019). Automated Kurdish Text Normalization. In *The Second International Conference on Kurdish and Persian Languages and Literature*. Sanandaj, Iran.

Mohammadamini, M., Veisi, H., Mahmudi, A., & Hosseini, H. (2019). Challenges in Standardization of Kurdish Language: A Corpus based approach. In *The Second International Conference on Kurdish and Persian Languages and Literature*. Sanandaj.

Nebez, J. (1993). The Kurdish Language From Oral Tradition to Written Language. In *Conference of "The Kurdish language toward the year 2000", organized by Sorbonne University and the Kurdish Institute in Paris*. Paris: London: Western Kurdistan Association publications.

Povey, D., Ghoshal, A., Boulianne, G., Burget, L., Glembek, O., Goel, N., et al. (2011). The Kaldi speech recognition toolkit. In *IEEE 2011 workshop on automatic speech recognition and understanding*. IEEE Signal Processing Society.

Sakoe, H., & Chiba, S. (1978). Dynamic programming algorithm optimization for spoken word recognition. *IEEE transactions on acoustics, speech, and signal processing*, *26*(1), 43–49.

Sheykh Esmaili, K., Eliassi, D., Salavati, S., Aliabadi, P., Mohammadi, A., Yosefi, S., & Hakimi, S. (2013). Building a Test Collection for Sorani Kurdish. *Proceedings of the 10th IEEE/ACS International Conference on Computer Systems and Applications, AICCSA*. https://doi.org/10.1109/AICCSA.2013.6616470

Sheykh Esmaili, K., & Salavati, S. (2013). Sorani Kurdish versus Kurmanji Kurdish: an empirical comparison. In *Proceedings of the 51st Annual Meeting of the Association for Computational Linguistics (Volume 2: Short Papers)* (pp. 300–305).

The Kurdish Academy. (2010). Rasipardekanî Konfransî berew Rênûsêkî Yekgrtûy Kurdî. *The Journal of Kurdish Academy*, (16), 14–16.

Veisi, H., MohammadAmini, M., & Hosseini, H. (2019). Toward Kurdish language processing: Experiments in collecting and processing the AsoSoft text corpus. *Digital Scholarship in the Humanities*. https://doi.org/10.1093/llc/fqy074

Vintsyuk, T. K. (1968). Speech discrimination by dynamic programming. *Cybernetics*, *4*(1), 52–57.

Walther, G., & Sagot, B. (2010). Developing a Large-Scale Lexicon for a Less-Resourced Language : General Methodology and Preliminary Experiments on Sorani Kurdish. *Proceedings of the 7th SaLTMiL Workshop on Creation and use of basic lexical resources for less-resourced languages (LREC 2010 Workshop)*.

Walther, G., Sagot, B., & Fort, K. (2010). Fast development of basic NLP tools: Towards a lexicon and a POS tagger for Kurmanji Kurdish. In *International conference on lexis and grammar*. Belgrade, Serbia.

Yu, D., & Deng, L. (2016). *Automatic Speech Recognition*. Springer.